\documentclass[11pt]{article}
\usepackage{colacl}
\newcommand{\exnum}[2]{\addtocounter{examplectr}{#1}(\arabic{examplectr}{#2})\addtocounter{examplectr}{-#1}}
\newcommand{\etal}{{\it et al.}}
\newcommand{\dsp}{\setlength{\baselineskip}{12.2pt}}
\newcommand{\ssp}{\setlength{\baselineskip}{10.5pt}}
\newcounter{examplectr}
\newcounter{subexamplectr}
\newenvironment{ex}%
   {\vspace{.1in}\addtocounter{examplectr}{1}
     \setcounter{subexamplectr}{0}
     \begin{list}
       {(\arabic{examplectr})}%
       {\setlength{\topsep}{0in}
        \setlength{\leftmargin}{.25in}
               \setlength{\labelsep}{0.075in}}
       \item \begin{minipage}[t]{2.8in} 
   }%
   {\end{minipage}
    \end{list}\vspace{.1in}}
\newenvironment{subex}%
   { \addtocounter{subexamplectr}{1}
     \begin{list}
       {\alph{subexamplectr}}%
       {\setlength{\topsep}{-\parskip}
        \setlength{\leftmargin}{0.175in}
        \setlength{\labelsep}{0.075in}}
       \item
   }%
   {\end{list}}
\title{\vspace{-2mm}{\it\small In Proceedings of the 6th ACL/SIGDAT
Workshop on Very Large Corpora, Montreal, Canada, 1998}\\[2.5mm]
Can Subcategorisation Probabilities Help a Statistical Parser?\\[1mm]}
\author{John Carroll \and Guido Minnen \\[0.5mm]
  School of Cognitive and Computing Sciences \\
  University of Sussex, Brighton, BN1 9QH, UK \\
  {\tt \{johnca,guidomi\}@cogs.susx.ac.uk}
	 \AND
	 Ted Briscoe \\[0.3mm]
  Computer Laboratory, University of Cambridge \\
  Pembroke Street, Cambridge CB2 3QG, UK \\
  {\tt ejb@cl.cam.ac.uk}}
\begin{document}
\dsp
\maketitle
\begin{abstract}
Research into the automatic acquisition of lexical information from corpora
is starting to produce large-scale computational lexicons containing data
on the relative frequencies of subcategorisation alternatives for
individual verbal predicates. However, the empirical question of whether
this type of frequency information can in practice improve the accuracy of
a statistical parser has not yet been answered. In this paper we describe an
experiment with a wide-coverage statistical grammar and parser for English
and subcategorisation frequencies acquired from ten million words of text
which shows that this information can significantly improve parse
accuracy\footnote{This work was funded by UK EPSRC project GR/L53175
`PSET: Practical Simplification of English Text', CEC Telematics
Applications Programme project LE1-2111 `SPARKLE: Shallow PARsing and
Knowledge extraction for Language Engineering', and by an EPSRC Advanced
Fellowship to the first author. Some of the work was carried out while the
first author was a visitor at the Tanaka Laboratory, Department of
Computer Science, Tokyo Institute of Technology, and at CSLI, Stanford
University; the author wishes to thank researchers at these institutions for
many stimulating conversations.}.
\end{abstract}

\section{Introduction}

Recent work on the automatic acquisition of lexical information from
substantial amounts of machine-readable text (e.g.\ Briscoe \& Carroll, 1997;
Gahl, 1998; Carroll \& Rooth, 1998) has opened up the possibility of
producing large-scale computational lexicons containing data on the relative
frequencies of subcategorisation alternatives for individual verbal
predicates. However, although Resnik (1992), Schabes (1992), Carroll \& Weir
(1997) and others have proposed `lexicalised' probabilistic grammars to
improve the accuracy of parse ranking, no wide-coverage parser has yet been
constructed which explicitly incorporates probabilities of different
subcategorisation alternatives for individual predicates. It is therefore an
open question whether this type of information can actually improve parser
accuracy in practice.

In this paper we address this issue, describing an
experiment with an existing wide-coverage statistical grammar and parser for
English (Carroll \& Briscoe, 1996) in conjunction with subcategorisation
frequencies acquired from 10 million words of text from the British
National Corpus (BNC; Leech, 1992). Our results show conclusively that this
information can improve parse accuracy.

\section{Background}

\subsection{Subcategorisation Acquisition}

Several substantial machine-readable subcategorisation dictionaries
exist for English, either built semi-automatically from
machine-readable versions of conventional learners' dictionaries, or
manually by (computational) linguists (e.g.\ the Alvey NL Tools (ANLT)
dictionary, Boguraev \etal\ (1987); the COMLEX Syntax dictionary,
Grishman, Macleod \& Meyers (1994)). However, since these
efforts were not carried out in tandem with rigorous large-scale
classification of corpus data, none of the resources produced provide useful
information on the relative frequency of different subcategorisation frames.

Systems which are able to acquire a small number of verbal subcategorisation
classes automatically from corpus text have been described by Brent (1991,
1993), and Ushioda \etal\ (1993). Ushioda
\etal\ also derive relative subcategorisation frequency information for
individual predicates. In this work they utilise a part-of-speech (PoS)
tagged corpus and finite-state NP parser to recognise and calculate the
relative frequency of six subcategorisation classes. They report that for 32
out of 33 verbs tested their system correctly predicts the most frequent
class, and for 30 verbs it correctly predicts the second most frequent
class, if there was one.

Manning (1993) reports a larger experiment, also using a PoS tagged
corpus and a finite-state NP parser, attempting to recognise sixteen
distinct complementation patterns---although not with relative frequencies.
In a comparison
between entries for 40 common verbs acquired from 4.1 million words of
text and the entries given in the {\it Oxford Advanced Learner's Dictionary
of Current English} (Hornby, 1989) Manning's system achieves a precision
of 90\% and a recall of 43\%.

Gahl (1998) presents an extraction tool for use with the BNC that is
able to create subcorpora containing different subcategorisation frames for
verbs, nouns and adjectives, given the frames expected for each predicate. The
tool is based on a set of regular expressions over PoS tags, lemmas,
morphosyntactic tags and sentence boundaries, effectively performing the same
function as a chunking parser (c.f.\ Abney, 1996). The resulting subcorpora
can be used to determine the (relative) frequencies of the frames.

Carroll \& Rooth (1998) use an iterative approach to estimate the
distribution of subcategorisation frames given head words, starting from a
manually-developed context-free grammar (of English). First, a probabilistic
version of the grammar is trained from a text corpus using the
expectation-maximisation (EM) algorithm, and the grammar is lexicalised
on rule heads. The EM algorithm is then run again to calculate the
expected frequencies of a head word accompanied by a particular frame. These
probabilities can then be fed back into the grammar for the next iteration.
Carroll \& Rooth report encouraging results for three verbs based on
applying the technique to text from the BNC. 

Briscoe \& Carroll (1997) describe a system capable of distinguishing 160
verbal subcategorisation classes---a superset of those found in the ANLT and
COMLEX Syntax dictionaries---returning relative frequencies for each frame
found for each verb. The classes also incorporate information about control
of predicative arguments and alternations such as particle movement and
extraposition. The approach uses a robust statistical parser which yields
complete though `shallow' parses, a comprehensive subcategorisation class
classifier, and {\it a priori} estimates of the probability of membership of
these classes. For a sample of seven verbs with multiple subcategorisation
possibilities the system's frequency rankings averaged 81\% correct. (We 
talk about this system further in section~\ref{sec-subcat} below, describing
how we used it to provide frequency data for our experiment).

\subsection{Lexicalised Statistical Parsing}

Carroll \& Weir (1997)---without actually building a parsing system---address
the issue of how frequency information can be associated with lexicalised
grammar formalisms, using Lexicalized Tree Adjoining Grammar (Joshi \&
Schabes, 1991) as a unifying framework. They consider systematically a
number of alternative probabilistic formulations, including those of Resnik
(1992) and Schabes (1992) and implemented systems based on other underlying
grammatical frameworks, evaluating their adequacy from both a theoretical
and empirical perspective in terms of their ability to model particular
distributions of data that occur in existing treebanks.

Magerman (1995), Collins (1996), Ratnaparkhi (1997), Charniak (1997) and
others describe implemented systems with impressive accuracy on parsing
unseen data from the Penn Treebank (Marcus, Santorini \&
Marc\-in\-kie\-wicz, 1993). These parsers model probabilistically the
strengths of association between heads of phrases, and the configurations in
which these lexical associations occur. The accuracies reported for these
systems are substantially better than their (non-lexicalised) probabilistic
context-free grammar analogues, demonstrating clearly the value of
lexico-statistical information. However, since the grammatical
descriptions are induced from atomic-labeled constituent structures in
the training treebank, rather than coming from an explicit generative
grammar, these systems do not make contact with traditional notions of
argument structure (i.e.\ subcategorisation, selectional preferences of
predicates for complements) in any direct sense. So although it is now
possible to extract at least subcategorisation data from large
corpora\footnote{Grishman
\& Sterling (1992), Poznanski \& Sanfilippo (1993), Resnik (1993), Ribas
(1994), McCarthy (1997) and others have shown that it is possible also to
acquire selection preferences automatically from (partially) parsed data.}
with some degree of reliability, it would be difficult to integrate the data
into this type of parsing system.

Briscoe \& Carroll (1997) present a small-scale
experiment in which subcategorisation class frequency
information for individual verbs was integrated into a robust statistical
(non-lexicalis\-ed) parser. The experiment used a test corpus of 250
sentences, and used the standard GEIG bracket precision, recall and crossing
measures (Grishman, Macleod \& Sterling, 1992) for evaluation. While bracket
precision and recall were virtually unchanged, the crossing bracket score
for the lexicalised parser showed a 7\% improvement. However, this
difference turned out not to be statistically significant at the 95\% level:
some analyses got better while others got worse.

We have performed a similar, but much larger scale experiment, which we
describe below. We used a larger test
corpus, acquired data from an acquisition corpus an order of
magnitude larger, and used a different quantitative evaluation measure that
we argue is more sensitive to argument/adjunct and attachment distinctions.
We summarise the main features of the `baseline' parsing system in
section~\ref{sec-baseline}, describe how we lexicalised it
(section~\ref{sec-subcat}), present the results of the quantitative
evaluation (section~\ref{sec-eval}), give a qualitative analysis of the
analysis errors made (section~\ref{sec-discuss}), and conclude with
directions for future work.

\section{The Experiment}

\subsection{The Baseline Parser}\label{sec-baseline}

The baseline parsing system comprises: 
\begin{itemize}
\item an HMM part-of-speech tagger (Elworthy, 1994), which produces either
the single highest-ranked tag for each word, or multiple tags with associated
forward-backward probabilities (which are used with a threshold to prune
lexical ambiguity);
\item a robust finite-state lemmatiser for English, an
extended and enhanced version of the University of Sheffield GATE system
morphological analyser (Cunningham \etal, 1995);
\item a wide-coverage unification-based `phrasal' grammar of English
PoS tags and punctuation;
\item a fast generalised LR parser using this grammar, taking the
results of the tagger as input, and performing disambiguation using a
probabilistic model similar to that of Briscoe \& Carroll (1993); and
\item training and test treebanks (of 4600 and 500 sentences
respectively) derived semi-automat\-ic\-ally from the {\sc susanne} corpus
(Sampson, 1995);
\end{itemize}

The grammar consists of 455 phrase structure rule schemata in the
format accepted by the parser (a syntactic variant of a Definite
Clause Grammar with iterative (Kleene) operators). It is `shallow' in
that no attempt is made to fully analyse unbounded dependencies.
However, the distinction between arguments and adjuncts is expressed,
following X-bar theory, by Chomsky-adjunction
to maximal projections of adjuncts (\mbox{$XP \rightarrow XP\ 
Adjunct$}) as opposed to `government' of arguments (i.e.\ arguments are
sisters within X1 projections; \mbox{$X1 \rightarrow X0\ Arg1\ ...\ 
ArgN$}). Furthermore, all analyses are rooted (in $S$) so the grammar
assigns global, shallow and often `spurious' analyses to many
sentences. Currently, the
coverage of this grammar---the proportion of sentences for which at least one
analysis is found---is 79\% when applied to the {\sc susanne} corpus, a 138K
word treebanked and balanced subset of the Brown corpus.

Inui \etal\ (1997) have recently proposed a novel model for probabilistic LR
parsing which they justify as theoretically more consistent and principled
than the Briscoe \& Carroll (1993) model. We use this new model since we have
found that it indeed also improves disambiguation accuracy.

The 500-sentence test corpus consists only of in-coverage sentences, and
contains a mix of written genres: news reportage (general and sports), {\it
belles lettres}, biography, memoirs, and scientific writing. The mean
sentence length is 19.3 words (including punctuation tokens).

\subsection{Incorporating Acquired Subcategorisation
Information}\label{sec-subcat}

The test corpus contains a total of 485 distinct verb lemmas. We ran the
Briscoe \& Carroll (1997) subcategorisation acquisition system on the first
10 million words of the BNC, for each of
these verbs saving the first 1000 cases in which a possible
instance of a subcategorisation frame was identified. For each
verb the acquisition system hypothesised a set of lexical entries
corresponding to frames for which it found enough evidence. Over the
complete set of verbs we ended up with a total of 5228 entries, each with an
associated frequency normalised with respect to the total number of frames
for all hypothesised entries for the particular verb. 

In the experiment each acquired lexical entry was assigned a probability
based on its normalised frequency, with smoothing---to allow for
unseen events---using the (comparatively crude) {\it add-1} technique. We
did not use the lexical entries themselves during parsing, since missing
entries would have compromised coverage. Instead, we factored in their
probabilities during parse ranking at the end of the parsing process.

We ranked complete derivations based on the product of (1)~the (purely
structural) derivation probability according to the probabilistic LR model,
and (2)~for each verb instance in the derivation the
probability of the verbal lexical entry that would be used in the particular
analysis context. The entry was located via the {\it VSUBCAT} value
assigned to the verb in the analysis by the immediately dominating verbal
phrase structure rule in the grammar: {\it VSUBCAT} values are also present
in the lexical entries since they were acquired using the same grammar.
Table~\ref{classes-used} lists the {\it VSUBCAT} values.
\begin{table}
\begin{center}
{\small\tt
\begin{tabular}{|llll|}\hline
AP & NP\_PP\_PP & PP\_WHPP & VPINF\\
NONE & NP\_SCOMP & PP\_WHS & VPING\\
NP & NP\_WHPP & PP\_WHVP & VPING\_PP\\
NP\_AP & PP & SCOMP & VPPRT\\
NP\_NP & PP\_AP & SINF & WHPP\\
NP\_NP\_SCOMP & PP\_PP & SING &\\
NP\_PP & PP\_SCOMP & SING\_PP &\\
NP\_PPOF & PP\_VPINF & VPBSE &\\ \hline
\end{tabular}}
\end{center}
\caption{{\it VSUBCAT} values in the grammar.}
\label{classes-used}
\end{table}
The values are mostly self-explanatory; however, examples of
some of the less obvious ones are given in \exnum{+1}{}.
\begin{ex}
{\it
 They made {\small\tt (NP\_WHPP)} a great fuss about what to do.\\
 They admitted {\small\tt (PP\_SCOMP)} to the authorities that they had
 entered illegally.\\
 It dawned {\small\tt (PP\_WHS)} on him what he should do.
}
\end{ex}
Some {\it VSUBCAT} values correspond to several of the 160
subcategorisation classes distinguished by the acquisition system. In
these cases the sum of the probabilities of the corresponding entries was
used. The finer distinctions stem from the use by the acquisition system of
additional information about classes of specific prepositions, particles and
other function words appearing within verbal frames. In this experiment we
ignored these distinctions. 

In taking the product of the derivation and subcategorisation probabilities
we have lost some of the properties of a statistical language model. The
product is no longer strictly a probability, although we do not attempt to
use it as such: we use it merely to rank competing analyses. Better
integration of these two sets of probabilities is an area which requires
further investigation.

\subsection{Quantitative Evaluation}\label{sec-eval}

\subsubsection{Bracketing}

We evaluated parser accuracy on the unseen test corpus with respect to the
phrasal bracketing annotation standard described by Carroll \etal\ (1997)
rather than the original {\sc susanne} bracketings, since the analyses
assigned by the grammar and by the corpus differ for many
constructions\footnote{Our previous attempts to produce {\sc susanne}
annotation scheme analyses were not entirely successful, since {\sc
susanne} does not have an underlying grammar, or even a formal
description of the possible bracketing configurations. Our evaluation
results were often more sensitive to the exact mapping we used than to
changes we made to the parsing system itself.}. However, with the exception
of {\sc susanne} `verb groups' our annotation standard is bracket-consistent
with the treebank analyses (i.e.\ no `crossing brackets'). 
Table~\ref{bracket-lexicalised} shows the baseline accuracy of the parser
with respect to (unlabelled) bracketings, and also with this model when
augmented with the extracted subcategorisation information.
\begin{table*}
\begin{center}
\begin{tabular}{|l|rrrr|} \hline
   & \multicolumn{1}{c}{Zero} & \multicolumn{1}{c}{Mean} &
\multicolumn{1}{c}{Bracket} & \multicolumn{1}{c|}{Bracket} \\

   & \multicolumn{1}{c}{crossings} & \multicolumn{1}{c}{crossings} &
\multicolumn{1}{c}{recall} &  \multicolumn{1}{c|}{precision} \\

   & \multicolumn{1}{c}{(\% sents.)} & \multicolumn{1}{c}{per sent.} &
\multicolumn{1}{c}{(\%)} & \multicolumn{1}{c|}{(\%)} \\
\hline

`Baseline'
&      57.2     & 1.11             & 82.5             & 83.0 \\[0.5mm]
With subcat
&      56.6     & 1.10             & 83.1             & 83.1 \\ \hline
\end{tabular}

\caption{Bracketing evaluation measures, before and after
incorporation of subcat information}
\label{bracket-lexicalised}
\end{center}
\end{table*}
Briefly, the evaluation metrics compare unlabelled bracketings derived from
the test treebank with those derived from parses, computing {\it recall},
the ratio of matched brackets over all brackets in the treebank; {\it
precision}, the ratio of matched brackets over all brackets found by the
parser; {\it mean crossings}, the number of times a bracketed sequence
output by the parser overlaps with one from the treebank but neither is
properly contained in the other, averaged over all sentences; and {\it zero
crossings}, the percentage of sentences for which the analysis returned has
zero crossings (see Grishman, Macleod \& Sterling, 1992).

Since the test corpus contains only in-coverage sentences our results are
relative to the 80\% or so of sentences that can be parsed. In experiments
measuring the coverage of our system (Carroll \& Briscoe, 1996), we found
that the mean length of failing sentences was little different to that
of successfully parsed ones. We would therefore argue that
the remaining 20\% of sentences are not significantly more complex, and
therefore our results are not skewed due to parse failures. 
Indeed, in these experiments a fair proportion of unsuccessfully parsed
sentences were elliptical noun or prepositional phrases, fragments from
dialogue and so forth, which we do not attempt to cover.

On these measures, there is no significant difference between the baseline
and lexicalised versions of the parser. In particular, the mean crossing
rates per sentence are almost identical. This is in spite of the fact that
the two versions return different highest-ranked analyses for 30\% of the
sentences in the test corpus. The reason for the similarity in scores
appears to be that the annotation scheme and evaluation measures are
relatively insensitive to argument/adjunct and attachment distinctions. For
example, in the sentence \exnum{+1}{} from the test corpus
\begin{ex}
{\it Salem ( AP ) -- the statewide meeting of war
mothers Tuesday in Salem will hear a greeting from Gov.\ Mark Hatfield.}
\end{ex}
the phrasal analyses returned by the baseline and lexicalised parsers are,
respectively \exnum{+1}{a} and \exnum{+1}{b}.
\begin{ex}
\begin{subex}
{\it ...\ (VP will hear (NP a greeting) (PP from (NP Gov.\ Mark Hatfield)))
...}
\end{subex}
\begin{subex}
{\it ...\ (VP will hear (NP a greeting (PP from (NP Gov.\ Mark Hatfield))))
...}
\end{subex}
\end{ex}
The latter is correct, but the former, incorrectly taking the PP to be an
argument of the verb, is penalised only lightly by the evaluation measures:
it has zero crossings, and 75\% recall and precision. This type of
annotation and evaluation scheme may be appropriate for a {\it phrasal}
parser, such as the baseline version of the parser, which does not have the
knowledge to resolve such ambiguities. Unfortunately, it masks differences
between such a phrasal parser and one which can use lexical information to
make informed decisions between complementation and modification
possibilities\footnote{Shortcomings of this combination of annotation and
evaluation scheme have been noted previously by Lin (1996), Carpenter \&
Manning (1997) and others. Carroll, Briscoe \& Sanfilippo (1998) summarise
the various criticisms that have been made.}.

\subsubsection{Grammatical Relation}

We therefore also evaluated the baseline and lexicalised parser against the
500 test sentences marked up in accordance with a second, grammatical
relation-based (GR) annotation scheme (described in detail by Carroll,
Briscoe \& Sanfilippo, 1998).

\begin{figure*}
\centering
{\tt    \setlength{\unitlength}{0.65pt}
\begin{picture}(480,230)
\thinlines    \put(354,88){\line(-3,-1){66}}
              \put(354,88){\line(3,-1){72}}

              \put(436,50){\line(3,-2){36}}
              \put(436,50){\line(-3,-2){36}}

              \put(289,50){\line(2,-1){47}}
              \put(289,50){\line(0,-1){24}}
              \put(289,50){\line(-2,-1){47}}

              \put(142,89){\line(2,-1){47}}
              \put(142,89){\line(0,-1){24}}
              \put(142,89){\line(-2,-1){47}}

              \put(270,166){\line(4,-3){82}}
              \put(270,166){\line(-1,-2){15}}
              \put(270,166){\line(-2,-1){120}}

              \put(251,121){\line(2,-3){37}}
              \put(251,121){\line(-6,-1){100}}

              \put(80,168){\line(5,-2){55}}
              \put(80,168){\line(0,-1){24}}
              \put(80,168){\line(-5,-2){55}}

              \put(176,210){\line(4,-1){93}}
              \put(176,210){\line(0,-1){24}}
              \put(176,210){\line(-4,-1){93}}

              \put(150,218){\small\it dependent}

              \put(70,174){\small\it mod}
              \put(150,174){arg\_mod}
              \put(260,174){\small\it arg}

              \put(218,126){\small\it subj\_or\_dobj}

              \put(10,134){\small\it ncmod}
              \put(67,134){\small\it xmod}
              \put(119,134){\small\it cmod}

              \put(129,94){subj}
              \put(340,94){\small\it comp}

              \put(65,54){ncsubj}
              \put(124,54){xsubj}
              \put(175,54){csubj}

              \put(280,54){\small\it obj}
              \put(411,54){clausal}

              \put(230,14){dobj}
              \put(277,14){obj2}
              \put(322,14){iobj}

              \put(385,14){xcomp}
              \put(455,14){ccomp}
\end{picture}}
\caption{Portions of GR hierarchy used. (Relations in {\it
italics} are not returned by the parser).}
\label{grs-used}
\end{figure*}
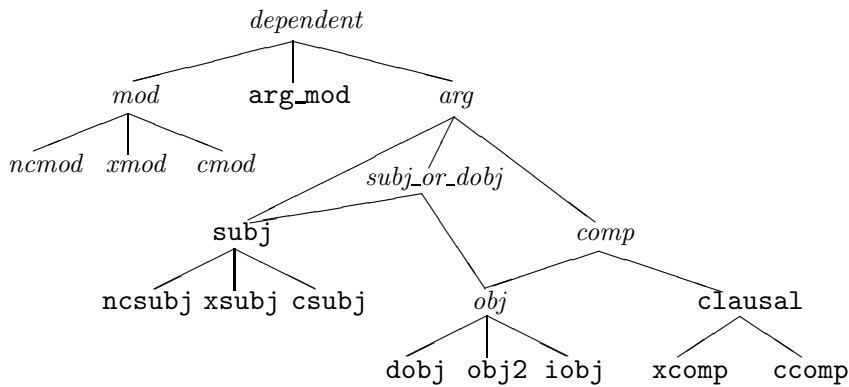

In general, grammatical relations (GRs) are viewed as specifying the
syntactic dependency which holds between a head and a dependent. The set of
GRs form a hierarchy; the ones we are concerned with are shown in
figure~\ref{grs-used}.
{\it Subj}\,(ect) GRs divide into clausal ({\it xsubj/csubj}\,), and
non-clausal ({\it ncsubj}\,) relations. {\it Comp}\,(lement) GRs divide into
{\it clausal}, and into non-clausal direct object ({\it dobj}\,), second
(non-clausal) complement in ditransitive constructions ({\it obj2}\,), and
indirect object complement introduced by a preposition ({\it
iobj}\,). 
In general the parser returns the most specific (leaf) relations in the GR
hierarchy, except when it is unable to determine whether clausal
subjects/objects are controlled from within or without (i.e.\ {\it csubj}
vs.\ {\it xsubj}, and {\it ccomp} vs.\ {\it xcomp} respectively), in which
case it returns {\it subj} or {\it clausal} as appropriate. Each relation is
parameterised with a head (lemma) and a dependent (lemma)---also optionally
a type and/or specification of grammatical function.
For example, the sentence \exnum{+1}{a} would be marked up as in
\exnum{+1}{b}.
\begin{ex}
\begin{subex}
 {\it Paul intends to leave IBM.}
\end{subex}
\begin{subex}
 {\it ncsubj\,(intend,Paul,\_)\\
 xcomp\,(to,intend,leave)\\
 ncsubj\,(leave,Paul,\_)\\
 dobj\,(leave,IBM,\_)}
\end{subex}
\end{ex}
Carroll, Briscoe \& Sanfilippo (1998) justify this new evaluation annotation
scheme and compare it with others (constituent- and dependency-based) that
have been proposed in the literature.

The relatively large size of the test corpus has meant that to date we
have in some cases not distinguished between {\it c/xsubj} and between {\it
c/xcomp}, and we have not marked up modification relations; we thus report
evaluation with respect to argument relations only (but including the
relation {\it arg\_mod}---a semantic argument which is syntactically
realised as a modifier, such as the passive `by-phrase'). The mean number of
GRs per sentence in the test corpus is 4.15.

When computing matches between the GRs produced by the parser and those in
the corpus annotation, we allow a single level of subsumption: a
relation from the parser may be one level higher in the GR
hierarchy than the actual correct relation. For example, if the parser
returns {\it clausal}, this is taken to match both the more specific {\it
xcomp} and {\it ccomp}. Also, an unspecified filler ({\it \_}) for the type
slot in the {\it iobj} and {\it clausal} relations successfully matches any
actual specified filler. The head slot fillers are in all cases the base
forms of single head words, so for example, `multi-component' heads, such as
the names of people, places or organisations are reduced to one word; thus
the slot filler corresponding to {\it Mr.\ Bill Clinton} would be {\it
Clinton}. For real-world applications this might not be the desired
behaviour---one might instead want the token {\it Mr.\_Bill\_Clinton}. This
could be achieved by invoking a processing phase similar to the conventional
`named entity' identification task in information extraction.

Considering the previous example \exnum{-2}{}, but this time with respect to
GRs, the sets returned by the baseline and lexicalised parsers are
\exnum{+1}{a} and \exnum{+1}{b}, respectively.
\begin{ex}
\begin{subex}
 {\it
 ncsubj\,(hear,meeting,\_)\\
 dobj\,(hear,greeting,\_)\\
 iobj\,(from,hear,Hatfield)}
\end{subex}
\begin{subex}
 {\it
 ncsubj\,(hear,meeting,\_)\\
 dobj\,(hear,greeting,\_)}
\end{subex}
\end{ex}
The latter is correct, but the former, incorrectly taking the PP to be an
argument of the verb, {\it hear}, is penalised more heavily than in the
bracketing annotation and evaluation schemes: it gets only 67\% recall.
There is also no misleadingly low crossing score since there is no analogue
to this in the GR scheme.

Table~\ref{grs-lexicalised} gives the result of evaluating the baseline
and lexicalised versions of the parser on the GR annotation. 
\begin{table}
\begin{center}
\begin{tabular}{|l|rr|} \hline
   & \multicolumn{1}{c}{Recall} & \multicolumn{1}{c|}{Precision} \\

   & \multicolumn{1}{c}{(\%)} & \multicolumn{1}{c|}{(\%)} \\
\hline

`Baseline'
&      88.6     & 79.2 \\[0.5mm]
With subcat
&      88.1     & 88.2 \\ \hline
\end{tabular}

\caption{GR evaluation measures, before
and after incorporation of subcategorisation information. Argument relations
only.}
\label{grs-lexicalised}
\end{center}
\end{table}
The measures compare the set of GRs in the annotated test corpus with those
returned by the parser, in terms of {\it recall}, the percentage of GRs
correctly found by the parser out of all those in the treebank; and {\it
precision}, the percentage of GRs returned by the parser that are actually
correct. In the evaluation, GR recall of the lexicalised parser drops by
0.5\% compared with the baseline, while precision increases by 9.0\%. The
drop in recall is not statistically significant at the 95\% level ({\it
paired~t-test}, 1.46, 499~$df$, $p>0.1$), whereas the increase in precision
is significant even at the 99.95\% level ({\it paired~t-test}, 5.14,
499~$df$, $p<0.001$).

Table~\ref{grs-comparison} gives the number of each type of GR returned by
the two models, compared with the correct numbers in the test corpus.
\begin{table*}
\begin{center}
\begin{tabular}{|l|rrrrrrrrrr|} \hline
   & \multicolumn{1}{c}{\it arg\_mod} & \multicolumn{1}{c}{\it ccomp}
   & \multicolumn{1}{c}{\it clausal} & \multicolumn{1}{c}{\it csubj}
   & \multicolumn{1}{c}{\it dobj} & \multicolumn{1}{c}{\it iobj}
   & \multicolumn{1}{c}{\it ncsubj} & \multicolumn{1}{c}{\it obj2}
   & \multicolumn{1}{c}{\it subj} & \multicolumn{1}{c|}{\it xcomp} \\[1mm]
\hline

 `Baseline'
 &   16 &   39 &  202 &  4 &  415 &  327 & 1054 & 53 & 14 &  202\\[0.5mm]
 With subcat
 &    9 &   20 &  138 &  3 &  429 &  172 & 1058 & 39 & 15 &  195\\[0.5mm]
 Correct
 &   32 &   16 &  136 &  2 &  428 &  160 & 1064 & 23 & 13 &  203\\ \hline
\end{tabular}

\caption{Numbers of each type of grammatical relation.}
\label{grs-comparison}
\end{center}
\end{table*}
The baseline parser returns
a mean of 4.65 relations per sentence, whereas the lexicalised
parser returns only 4.15, the same as the test corpus. This is
further, indirect evidence that the lexicalised probabilistic system models
the data more accurately.

\subsection{Discussion}\label{sec-discuss}

In addition to the quantitative analysis of parser accuracy reported above,
we have also performed a qualitative analysis of the errors made. We looked
at each of the errors made by the lexicalised version of the parser on the
500-sentence test corpus, and categorised them into errors concerning:
complementation, modification, co-ordination, structural attachment of
textual adjuncts, and phrase-internal misbracketing. Of course, multiple
errors within a given sentence may interact, in the sense that one error may
so disrupt the structure of an analysis that it necessarily leads to one or
more other errors being made. In all cases, though, we considered all of
the errors and did not attempt to determine whether or not one of them was
the `root cause'. Table~\ref{error-analysis} summarises the number of errors
of each type over the test corpus.
\begin{table}
\begin{center}
\begin{tabular}{|l|r|} \hline
      & Number \\ \hline

Complementation & 124 \\
Modification    & 134 \\
Co-ordination   & 30 \\
Textual         & 30 \\
Misbracketing   & 40 \\ \hline
\end{tabular}

\caption{Numbers of errors of each type made by the lexicalised parser.}
\label{error-analysis}
\end{center}
\end{table}

Typical examples of the five error types identified are:
\begin{description}
\item[complementation] {\it ...\ decried the high rate of unemployment in the
state} misanalysed as {\it decry} followed by an NP and a PP complement;
\item[modification] in {\it ...\ surveillance of the pricing practices of the
concessionaires for the purpose of keeping the prices reasonable}, the PP
modifier {\it for the purpose of ...} attached `low' to {\it
concessionaires} rather than `high' to {\it surveillance};
\item[co-ordination] the NP {\it priests, soldiers, and other members of the
party} misanalysed as just two conjuncts, with the first conjunct containing
the first two words in apposition;
\item[textual] in {\it But you want a job guaranteed when you return, I
continued my attack}, the (textual) adjunct {\it I ...\ attack} attached to
the VP {\it guaranteed ...\ return} rather than the S {\it But ...\ return};
and
\item[misbracketing] {\it Nowhere in Isfahan is this rich aesthetic life
of the Persians ...} has {\it of} misanalysed as a particle, with {\it
the Persians} becoming a separate NP.
\end{description}

There are no obvious trends within each type of error, although some
particularly numerous sub-types can be identified. In 8 of the 30 cases of
textual misanalysis, a sentential textual adjunct preceded by a comma was
attached too low. The most common type of modification error was---in 20 of
the 134 cases---misattachment of a $PP$ modifier of $\overline{N}$ to a
higher $VP$. The majority of the complementation errors were verbal,
accounting for 115 of the total of 124. In 15 cases of incorrect verbal
complementation a passive construction was incorrectly analysed as active,
often with a following `by' prepositional phrase erroneously taken to be a
complement.

Other shortcomings of the system were evident in the treatment of
co-ordinated verbal heads, and of phrasal verbs.  The grammatical relation
extraction module is currently unable to return GRs in which the verbal head
alone appears in the sentence as a conjunct---as in the VP {\it ...\ to
challenge and counter-challenge the authentication}. This can be remedied
fairly easily. Phrasal verbs, such as {\it to consist of} are identified as
such by the subcategorisation acquisition system. The grammar used by the
shallow parser analyses phrasal verbs in two stages: firstly the verb itself
and the following particle are combined to form a sub-constituent, and then
phrasal complements are attached. The simple mapping from {\it VSUBCAT}
values to subcategorisation classes cannot cope with the second
level of embedding of phrasal verbs, so these verbs do not pick up any
lexical information at parse time.

\section{Conclusions}

We surveyed recent work on automatic acquisition from corpora of
subcategorisation and associated frequency information. We described an
experiment with a wide-coverage statistical grammar and parser for English
and subcategorisation frequencies acquired from 10 million words of text
which shows that this information can significantly improve the accuracy of
recovery of grammatical relation specifications from a test corpus of 500
sentences covering a number or different genres.

Future work will include: investigating more principled probabilistic
models; addressing immediate lower-level shortcomings in the current system
as discussed in section~\ref{sec-discuss} above; adding {\it
mod}\,(ification) GR annotations to the test corpus and extending the parser
to also return these; and working on incorporating selectional preference
information that we are acquiring in other, related work (McCarthy, 1997).

\section*{References}

\newcommand{\book}[4]{\item #1 (#4). {\it #2}. #3.}
\newcommand{\barticle}[7]{\item #1 (#7). #2. In #5 (Eds.), {\it #4},
#3. #6.}
\newcommand{\bparticle}[6]{\item #1 (#6). #2. In #4 (Eds.), {\it #3}. #5.}
\newcommand{\boarticle}[5]{\item #1 (#5). #2. In {\it #3}. #4.}
\newcommand{\farticle}[6]{\item #1 (#6). #2. In #4 (Eds.), {\it #3}: #5.
Forthcoming.}
\newcommand{\uarticle}[5]{\item #1 (#5). #2. In #4 (Eds.), {\it #3}.
Forthcoming.}
\newcommand{\jarticle}[6]{\item #1 (#6). #2. {\it #3}, #4, #5.}
\newcommand{\particle}[6]{\item #1 (#6). #2. In {\it Proceedings of the
#3}, #5. #4.}
\newcommand{\lazyparticle}[5]{\item #1 (#5). #2. In {\it Proceedings of the
#3}. #4.}
\newcommand{\lazyjarticle}[4]{\item #1 (#4). #2. {\it #3}.}
\newcommand{\lazyfjarticle}[4]{\item #1 (#4). #2. {\it #3}. Forthcoming.}
\newcommand{\bookartnopp}[6]{\item #1 (#6) #2. In #4 (Eds.), {\it #3,} #5.}

\begin{list}{}
   {\leftmargin 0pt
    \itemindent 0pt
    \itemsep 1pt plus 1pt
    \parsep 1pt plus 1pt}

{\small
\ssp

\jarticle{Abney, S.}
{Partial parsing via finite-State cascades}
{Natural Language Engineering}
{2(4)}
{337--344}
{1996}

\particle{Boguraev, B., Briscoe, E., Carroll, J., Carter, D. \&
Grover, C.}
{The derivation of a grammatically-indexed lexicon from the Longman
Dictionary of Contemporary English}
{25th Annual Meeting of the Association for Computational Linguistics}
{Stanford, CA}
{193--200}
{1987}

\particle{Brent, M.}
         {Automatic acquisition of subcategorization frames from untagged text}
         {29th Annual Meeting of the Association for Computational Linguistics}
         {Berkeley, CA}
         {209--214}
         {1991}

\jarticle{Brent, M.}
         {From grammar to lexicon: unsupervised learning of lexical syntax}
         {Computational Linguistics}
         {19(3)}
         {243--262}
         {1993}

\jarticle{Briscoe, E. \& Carroll, J.}
     {Generalized probabilistic LR parsing for unification-based grammars}
     {Computational Linguistics}
     {19(1)}
     {25--60}
     {1993}

\lazyparticle{Briscoe, E. \& Carroll, J.}
{Automatic extraction of subcategorization from corpora}
{5th ACL Conference on Applied Natural Language Processing}
{Washington, DC}
{1997}

\lazyparticle{Carpenter, B. \& Manning, C.}
{Probabilistic parsing using left corner language models}
{5th ACL/SIGPARSE International Workshop on Parsing Technologies}  
{MIT, Cambridge, MA}
{1997}

\particle{Carroll, J. \& Briscoe, E.}
         {Apportioning development effort in a probabilistic LR parsing
system through evaluation}
         {1st ACL/SIGDAT Conference on Empirical Methods in Natural Language
Processing}
         {University of Pennsylvania, Philadelphia, PA}
         {92--100}
         {1996}

\book{Carroll, J., Briscoe, E., Calzolari, N., Federici, S.,
Montemagni, S., Pirrelli, V., Grefenstette, G., Sanfilippo, A.,
Carroll, G. \& Rooth, M.}
{SPARKLE WP1 specification of phrasal parsing}
{$<$http://www.ilc.pi.cnr.it/sparkle.html$>$}
{1997}

\particle{Carroll, J., Briscoe, E. \& Sanfilippo, A.}
{Parser evaluation: a survey and a new proposal}
{1st International Conference on Language Resources and Evaluation}
{Granada, Spain}
{447--454}
{1998}

\particle{Carroll, J. \& Weir, D.}
{Encoding frequency information in lexicalized grammars}
{5th ACL/SIGPARSE International Workshop on Parsing Technologies (IWPT-97)}
{MIT, Cambridge, MA}
{8--17}
{1997}

\lazyparticle{Carroll, G. \& Rooth, M.}
{Valence induction with a head-lexicalized PCFG}
{3rd Conference on Empirical Methods in Natural Language Processing}
{Granada, Spain}
{1998}

\particle{Charniak, E.}
{Statistical parsing with a context-free grammar and word statistics}
{14th National Conference on Artificial Intelligence (AAAI-97)} 
{Providence, RI} 
{598--603}
{1997}

\particle{Collins, M.}
{A new statistical parser based on bigram lexical dependencies}
{34th Meeting of the Association for Computational Linguistics}
{Santa Cruz, CA}
{184--191}
{1996}

\book{Cunningham, H., Gaizauskas, R. \& Wilks, Y.}
{A general architecture for text engineering (GATE) -- a new approach to
language R\&D}
{Research memo CS-95-21, Department of Computer Science, University of
Sheffield, UK}
{1995}

\lazyparticle{Elworthy, D.}
         {Does Baum-Welch re-estimation help taggers?}
         {4th ACL Conference on Applied Natural Language Processing}
         {Stuttgart, Germany}
         {1994}

\lazyparticle{Gahl, S.}
{Automatic extraction of subcorpora based on subcategorization frames
 from a part-of-speech tagged corpus}
{COLING-ACL'98}
{Montreal, Canada}
{1998}

\particle{Grishman, R., Macleod, C. \& Meyers, A.}
         {Comlex syntax: building a computational lexicon}
         {15th International Conference on Computational Linguistics
(COLING-94)}
         {Kyoto, Japan}
         {268--272}
         {1994}

\particle{Grishman, R., Macleod, C. \& Sterling, J.}
{Evaluating parsing strategies using standardized parse files}
{3rd ACL Conference on Applied Natural Language Processing}
{Trento, Italy}
{156--161}
{1992}

\particle{Grishman, R. \& Sterling, J.}
         {Acquisition of selectional patterns}
         {14th International Conference on Computational Linguistics
(COLING-92)}
         {Nantes, France}
         {658--664}
         {1992}

\book{Hornby, A.}
     {Oxford Advanced Learner's Dictionary of Current English}
     {Oxford, UK: OUP}
     {1989}

\particle{Inui, K., Sornlertlamvanich, V., Tanaka, H. \& Tokunaga, T.}
{A new formalization of probabilistic GLR parsing}
{5th ACL/SIGPARSE International Workshop on Parsing Technologies (IWPT-97)}
{Cambridge, MA}
{123--134}
{1997}

\bparticle{Joshi, A. \& Schabes, Y.}
{Tree-adjoining grammars and lexicalized grammars}
{Definability and Recognizability of Sets of Trees}
{M. Nivat \& A. Podelski}
{Elsevier}
{1991}

\jarticle{Leech, G.}
{100 million words of English: the British National Corpus}
{Language Research}
{28(1)}
{1--13}
{1992}

\barticle{Lin, D.}
{Dependency-based parser evaluation: a study with a software manual corpus}
{13--24}
{Industrial Parsing of Software Manuals}
{R. Sutcliffe, H-D. Koch \& A. McElligott}
{Amsterdam, The Netherlands: Rodopi}
{1996}

\lazyparticle{Magerman, D.}
          {Statistical decision-tree models for parsing}
          {33rd Annual Meeting of the Association for Computational Linguistics}
          {Boston, MA}
          {1995}

\particle{Manning, C.}
         {Automatic acquisition of a large subcategorisation
dictionary from corpora}
         {31st Annual Meeting of the Association for Computational Linguistics}
         {Columbus, Ohio}
         {235--242}
         {1993}

\jarticle{Marcus, M., Santorini, B. \& Marcinkiewicz}
{Building a large annotated corpus of English: The Penn Treebank}
{Computational Linguistics}
{19(2)}
{313--330}
{1993} 

\particle{McCarthy, D.}
{Word sense disambiguation for acquisition of selectional preferences}
{ACL/EACL'97 Workshop Automatic Information Extraction and Building of
Lexical Semantic Resources for NLP Applications}
{Madrid, Spain}
{52--61}
{1997}

\bparticle{Poznanski, V. \& Sanfilippo, A.}
         {Detecting dependencies between semantic verb subclasses and
subcategorization frames in text corpora}
         {SIGLEX ACL Workshop on the Acquisition of Lexical Knowledge from Text}
         {B. Boguraev \& J. Pustejovsky}
         {Columbus, Ohio}
         {1993}

\lazyparticle{Ratnaparkhi, A.}
{A linear observed time statistical parser based on maximum
entropy models}
{2nd Conference on Empirical Methods in Natural Language Processing}
{Brown University, Providence, RI}
{1997}

\particle{Resnik, P.}
{Probabilistic tree-adjoining grammar as a framework for statistical
natural language processing}
{14th International Conference on Computational Linguistics
  (COLING-92)}
{Nantes, France}
{418--424}
{1992}

\book{Resnik, P.}
     {Selection and information: a class-based approach to lexical 
relationships}
     {University of Pennsylvania, CIS Dept, PhD thesis}
     {1993}

\lazyparticle{Ribas, P.}
         {An experiment on learning appropriate selection restrictions from
a parsed corpus}
         {15th International Conference on Computational Linguistics
(COLING-94)}
         {Kyoto, Japan}
         {1994}

\book{Sampson, G.} {English for the computer} {Oxford, UK: Oxford University
Press} {1995}

\particle{Schabes, Y.}
{Stochastic lexicalized tree-adjoining grammars}
{14th International Conference on Computational Linguistics (COLING-92)}
{Nantes, France}
{426--432}
{1992}

\barticle{Ushioda, A., Evans, D., Gibson, T. \& Waibel, A.}
         {The automatic acquisition of frequencies of verb
subcategorization frames from tagged corpora}
         {95--106}
         {SIGLEX ACL Workshop on the Acquisition of Lexical Knowledge from Text}
         {B. Boguraev \& J. Pustejovsky}
         {Columbus, Ohio}
         {1993}

}
\end{list}

\end{document}